\documentclass[11pt,a4paper]{article}
\usepackage[hyperref]{acl2021}
\usepackage{cite,amsmath,graphicx,algorithmic,array,stfloats,color,times,latexsym,enumitem,booktabs,multirow,url,listings,amssymb,makecell,arydshln,longtable,soul}
\usepackage[caption=false,font=footnotesize]{subfig}
\usepackage[linesnumbered,ruled,vlined,algo2e]{algorithm2e}
\usepackage{microtype}
\usepackage{CJK}

\usepackage{arydshln}
\usepackage{colortbl}
\usepackage{microtype}
\aclfinalcopy 

\title{Self-Guided Curriculum Learning for Neural Machine Translation}

\author{First Author \\
  Affiliation / Address line 1 \\
  Affiliation / Address line 2 \\
  Affiliation / Address line 3 \\
  \texttt{email@domain} \\\And
  Second Author \\
  Affiliation / Address line 1 \\
  Affiliation / Address line 2 \\
  Affiliation / Address line 3 \\
  \texttt{email@domain} \\}

\author{
Lei Zhou\thanks{Partial of this work was done when the first author was visiting at CLSP, JHU.}\\
Nagoya University\\
\normalsize \textsf{zhou.lei@a.mbox.nagoya-u.ac.jp}\\
\And
Liang Ding\\
The University of Sydney\\
\normalsize \textsf{ldin3097@sydney.edu.au}\\
\And
Kevin Duh\\
Johns Hopkins University\\
\normalsize \textsf{kevinduh@cs.jhu.edu}\\
\AND
Shinji Watanabe\\
Carnegie Mellon University\\
\normalsize \textsf{shinjiw@ieee.org}\\
\And
Ryohei Sasano\\
Nagoya University\\
\normalsize \textsf{sasano@i.nagoya-u.ac.jp}\\
\And
Koichi Takeda\\
Nagoya University\\
\normalsize \textsf{takedasu@i.nagoya-u.ac.jp}\\
}
\date{}

\begin{document}
\maketitle
\begin{abstract}
In the field of machine learning, the well-trained model is assumed to be able to recover the training labels, i.e. the synthetic labels predicted by the model should be close to the ground-truth labels as much as possible. Inspired by this, we propose a self-guided curriculum strategy to encourage the learning of neural machine translation (NMT) models to follow the above recovery criterion, where we cast the recovery degree of each training example as its learning difficulty. Specifically, we adopt the sentence level BLEU score as the proxy of recovery degree. Different from existing curricula relying on linguistic prior knowledge or language models, 
our chosen learning difficulty is more suitable to measure the degree of knowledge mastery of the NMT models. Experiments on translation benchmarks, including WMT14 English$\Rightarrow$German and WMT17 Chinese$\Rightarrow$English, demonstrate that our approach can consistently improve translation performance against strong baseline Transformer. 
\end{abstract}

\section{Introduction}
\label{sec:intro}

Inspired by the learning behavior of human, Curriculum Learning (CL) for neural network training starts from a basic idea of starting small, namely better to start from easier aspects of a task and then progress towards aspects with increasing level of difficulty~\citep{elman1993learning}. \citet{bengio2009curriculum} achieves significant performance boost on tasks by forcing models to learn training examples following an order from "easy" to "difficult". They further explain CL method with two important constituents, how to rank training examples by learning difficulty, and how to schedule the presentation of training examples based on that rank.

\begin{figure}
\centering
\includegraphics[width=0.45\textwidth]{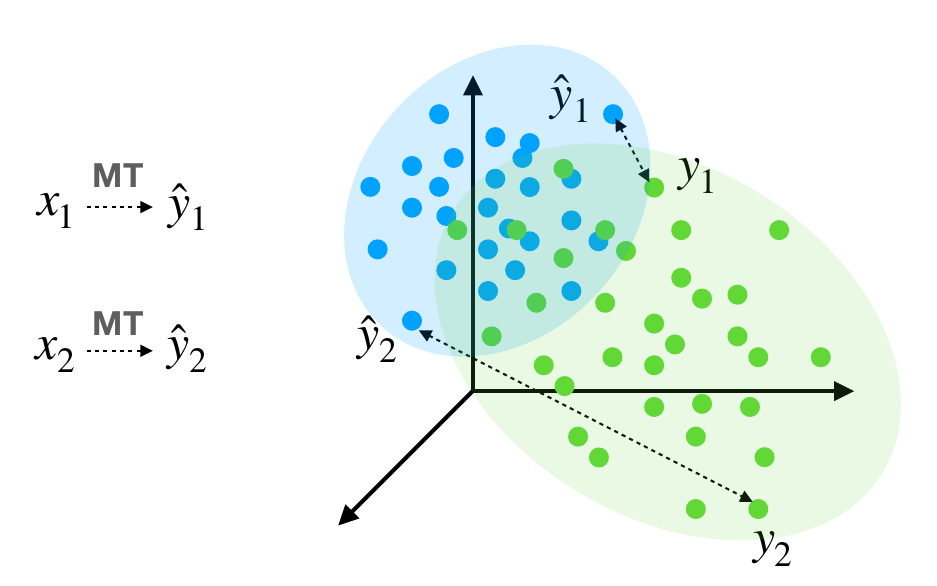}
\caption{\label{fig:mapping} The NMT model is well-trained on parallel corpus $\mathbb{D}$, $\{(x_1,y_1), (x_2,y_2)\} \in \mathbb{D}$. Taking $x_1$ and $x_2$ as the input, the \textit{recovery degrees} of $y_1$ are significantly better than that of $y_2$. Note that the distance between $y_i$ and $\hat{y_i}$ represents the recovery degrees, which indicate by dashed arrows.}
\end{figure}

In the scenario of neural machine translation (NMT), empirical studies have shown that CL strategy contributes to convergence speed and model performance~\citep{zhang2018empirical,platanios2019competence,zhang2019curriculum,liu2020norm,zhan2021meta,ruiter2020self}. In these approaches, the learning difficulty of each training example is measured by different difficulty criteria. Early approaches depend on prior knowledge from various sources, including manually crafted features like sentence length and word rarity~\citep{kocmi2017curriculum}. The drawback lies in the fact that human understand learning difficulty differently from NMT models. Recent works choose to derive difficulty criteria based on the probability distribution of training examples, to approximate the perspective of an NMT model. For example, ~\citet{platanios2019competence} turns discrete numerical difficulty scores into relative probabilities and then construct the criterion, while others derive criteria from independently pre-trained models like language model~\citep{zhang2019curriculum,dou2020dynamic,liu2020norm} and word embedding model~\citep{zhou2020uncertainty}. \citet{xu2020dynamic} derives criterion from the NMT model in the training process. And according to different way of scheduling the curriculum, these difficulty criteria are apply to either fixed schedule~\citep{cirik2016visualizing} or dynamic one~\citep{platanios2019competence, liu2020norm,xu2020dynamic,zhou2020uncertainty}.

 A well-trained NMT model learns an optimal probability distribution mapping sentences from source language to target language, which is expected to be capable of recovering training labels~\citep{liu2020understanding}. However if we test on the training set, we can observe inconsistent predictions against the target reference sentences, reflecting the discrepancy between the model distribution and the empirical distribution of training corpus, as Figure~\ref{fig:mapping} illustrated. For a training example, high recovery degree between prediction and target reference sentence means it's easier to be masted by the NMT model, while low recovery degree means it's more difficult~\citep{dingtao2019university,wu2020tencent}. Taking recovery degree as the difficulty criterion, we propose a CL strategy to schedule curriculum learning with a well-trained vanilla NMT model. We put forward an analogy of this method that a person can schedule a personal and effective curriculum after skimming over the whole textbook, namely \textit{self-guided curriculum}. 
 
 In this work, we cast recovery degree of each training example as its learning difficulty, enforcing an NMT model to learn from examples with higher recovery degree to the lower ones, and we analyze the coordination of this criterion with fixed and dynamic curriculum schedules. We conduct experiments on widely-used benchmarks, including WMT14 En-De and WMT17 Zh-En. Experimental results demonstrate that our proposed self-guided CL strategy can boost the performance of an NMT model against strong baseline Transformer.

\section{Problem Definition}
\label{sec:def}
For better interpretation of curriculum learning for neural machine translation, we put the discussion of various CL strategies into a probabilistic perspective. Such perspective also motivates us to derive this recovery degree based difficulty criterion.

\subsection{NMT Model}
Let $\mathcal{S}$ represent a probability distribution over all possible sequences of tokens in source language and $\mathcal{T}$ represent those over target language. Denote by $P_\mathcal{S}{(\mathbf{x})}$ the distribution of a random variable $\mathbf{x}$, each instance $x$ of which is a source sentence. Denote by $P_\mathcal{T}{(\mathbf{y})}$ the distribution of a random variable $\mathbf{y}$, each instance of which is a target sentence. NMT model is to learn a conditional distribution $P_{\mathcal{S},\mathcal{T}}{(\mathbf{y}|\mathbf{x})}$ with a probabilistic model $P(y|x;\theta)$ parameterized by $\theta$. And $\theta$ is learned by minimizing the objective:

\begin{equation}
    J(\theta) = -\mathbb{E}_{x,y \sim P_{\mathcal{S},\mathcal{T}}(\mathbf{x,y)}} \log P(y|x;\theta)
\end{equation}

\subsection{Curriculum Learning for Neural Machine Translation}
CL methods decompose the NMT model training process into a sequence of training phases, $1,\dots,K$, enforcing the optimization trajectory through the parameter space to visit a series of points ${\theta^{1},\dots,\theta^{K}}$ to minimize the objective $J(\theta)$. Given a parallel training corpus $\mathbb{D}$ and $K$ subsets $\{\mathbb{D}_1,\dots,\mathbb{D}_K\}  \subseteq \mathbb{D}$, each training phase can be viewed as a sub-optimal process trained on a subset:
\begin{equation}
\label{eq:nmt}
    J(\theta^{k+1}) = -\mathbb{E}_{x,y \sim \hat{P}_{\mathbb{D}_k}} \log P(y|x;\theta^k)
\end{equation}
where $\hat{P}_{\mathbb{D}_k}$ is the empirical distribution of $\mathbb{D}_k$. According to the definition of curriculum learning, the difficulty of ${J(\theta^1),\dots,J(\theta^K)}$ increases~\citep{bengio2009curriculum}. It is put into practice by grouping $\{\mathbb{D}_1,\dots,\mathbb{D}_K\}$ in an ascending order of learning difficulty. This process of splitting $\mathbb{D}$ into $K$ subsets can be formalized as follows:

\begin{itemize}
\item $\mathrm{score} \leftarrow d(z^n)$, $z^n \in \mathbb{D}$, where $d(\cdot)$ is a difficulty criterion
\item For $k=1,\dots,K$ do; $\mathbb{D}_k \leftarrow \{z^n | \mathrm{Constraint}(d(z^n),k)\}$
\end{itemize}

$z$ represents examples in $\mathbb{D}$, namely $\mathbb{D}=\{z^n\}_{n=1}^N$, $z^n=(x^n,y^n)$. Training corpus $\mathbb{D}$ is split into $K$ subsets $\{\mathbb{D}_1,\dots,\mathbb{D}_K\}$, that $\underset{k\in K}{\bigcup} \mathbb{D}_k=\mathbb{D}$.  

With these notations, we analyze the difficulty criteria in common CL strategies from a probabilistic perspective. As mentioned in Section~\ref{sec:intro}, except for numerical scores of manually crafted features, recent approaches generally derive their criteria from a probabilistic distribution. For example:

\paragraph{Explicit Feature} $d(x^n)=P_{\mathbb{D}}(\mathrm{Feature}(x^n))$, where $\mathrm{Feature}(\cdot)$ is handcrafted feature such as sentence length or word rarity. \citet{Ding2021UnderstandingAI} shows that explicit features, e.g. low-frequency words, may affect the model lexical choices, thus leading to different model performance. With the cumulative density function (CDF), numerical scores are mapped into a relative probability distribution over all training examples. Only features of source sentences are taken into consideration in~\citet{platanios2019competence}.

\paragraph{Language Model} $d(x^n)=-\frac{1}{I} \log P_{\mathrm{LM}}(w_1^n,\dots,w_I^n)$, where a language model pre-trained on source sentences of the parallel corpus, $\mathbb{D}_S$, is adopted to measure the uncertainly of each source sentence $x={w_1,\dots,w_i,\dots,w_I}$ by per-word cross-entropy~\citep{zhang2019curriculum}. $d(x)$ and $d(y)$ can be used separately or jointly. Both n-gram language model and neural language model are adopted in~\citet{zhou2020uncertainty}.

\paragraph{Word Embedding} $d(x^n)=\sum_{i=1}^I \Vert \mathbf{w}_i^n \Vert$, where $\mathbf{w}_1,\dots,\mathbf{w}_i,\dots,\mathbf{w}_I$ is a distributed representation of source sentence $x$ mapped through a independently trained word embedding model. In the case of~\citet{liu2020norm} the norm of word vector on the source side is used as the difficulty criterion. They also use the CDF function to assure the difficulty scores are within $[0,1]$.

\paragraph{NMT Model}
$d(z^n;\theta^k)=\frac{l(z^n;\theta^k)-l(z^n;\theta^{k-1})}{l(z^n;\theta^{k-1})}$,  $l(z^n;\theta^k)=-\log P(y^n|x^n;\theta^k)$, where $\theta^k$ represents the NMT model parameters at the $k$th training phase. The decline of loss is defined as the difficulty criterion in~\citet{xu2020dynamic}. Besides, the score of cross-lingual patterns may also be a proper difficulty criterion for NMT~\citep{ding2020self,zhou2020zero,wu2021slua}, which we leaves as future work. 

We now turn to \textit{curriculum scheduling}. There are two controlling factors, extraction of training set and training phase duration, namely how to split training corpus into subsets and when to load them. Given difficulty scores $d(z^n),z^n\in \mathbb{D}$, $\mathbb{D}$ is split into $K$ mutual exclusive subsets $\{\mathbb{D}_1,\dots,\mathbb{D}_K\}$, which are loaded in order as training phase progresses. There are two general regimens. In \textit{one pass} regimen, $k$ subsets $\mathbb{D}_k$ are loaded as training set one by one, while in \textit{baby steps} regimen, these subsets are merged into the current training set one by one~\citep{cirik2016visualizing}. According to~\citet{cirik2016visualizing}, baby steps outperforms one pass. Later approaches generally take the idea of baby steps in that easy examples are not cast aside while the probability increases for difficulty examples to be batched as training progresses. 

On top of baby steps, some works choose a \textit{fixed} setting. The size of training set is scaled up by a certain proportion of the total training examples, i.e. $|\mathbb{D}_k|=N/K$ when a new training phase begins. And each training phase spends a fixed number of training steps. Others choose a \textit{dynamic} setting. One is the \textit{competence} type~\citep{platanios2019competence,liu2020norm,xu2020dynamic}, where training set extraction happens during training. At the beginning of a training phase, it determines the upper limit of difficulty score by competence represented by $c(t)$ at $t$ steps, and all examples with difficulty scores lower than $c(t)$ will be extracted, namely $\{z^n | d(z^n) \leq c(t), z^n \in \mathbb{D}\}$, as the training set for current phase. With training set extraction being dynamic, the training duration is fixed. In competence-based scheduling, the range of difficulty scores ${d(z^n)}$ is $[0,1]$. Competence $c(t)$ is to determine $(K-1)$ upper limits within this range with a scale factor. A simple scale factor is training steps, $1,\dots,t,\dots,T$. With a initial value $c_0 \geqq 0$, the general form of competence function is : $c(t)=\min \left( 1,\sqrt[p]{t\frac{1-c^p_0}{T}+c^p_0} \right)$, where $p$ is the coefficient. When $p=1$, competence scale up linearly as training progresses. Larger $p$ means the number of training examples increases faster in early phases while slower in latter ones. Other scale factors can also be adoped, such as norm of source embeddings of an NMT model~\citep{liu2020norm} and BLEU score on validation set~\citep{xu2020dynamic}. Another one is the \textit{uncertainty} 
type~\citep{zhou2020uncertainty}, where training set extraction is fixed and happens before training. But the training duration, time steps spend on a training phase is controlled by a factor, i.e. model uncertainty, which is the variance of distribution over sampled examples with perturbed NMT model. Training process stays in a phase until model uncertainty stops decline.

\section{Methodology}


\begin{table*}
\renewcommand\arraystretch{1.2}
\centering
\begin{tabular}{ll}
\toprule
\multicolumn{2}{l}{\textbf{High Recovery Degree}   (BLEU 77.01)} \\
\hline
Source & {\begin{CJK}{UTF8}{gbsn}该动议如被通过, \textcolor{blue}{提案}或修正案中后被核准的各部分应合成整体再付表决。\end{CJK}} \\
\cdashline{1-2}
\multirow[t]{2}{*}{Reference} & If the motion for division is carried, those parts of \textcolor{blue}{the proposal} or of the amendment \\
& which are subsequently approved shall be put to the vote as a whole. \\
\cdashline{1-2}
\multirow[t]{2}{*}{Prediction} & 
If the motion for division is carried , those parts of \textcolor{red}{\ul{fm draft resolution}} or of the \\
& amendment that are subsequently approved shall be put to the vote as a whole. \\
\hline
\hline
\multicolumn{2}{l}{\textbf{Low Recovery Degree}   (BLEU 5.19)} \\
\hline
\multirow[t]{2}{*}{Source} & {\begin{CJK}{UTF8}{gbsn}并且慢慢地, 非常缓慢地把头\textcolor{blue}{抬}到它的眼睛正好\textcolor{blue}{可以直视}哈利的位置便停\end{CJK}}\\
& {\begin{CJK}{UTF8}{gbsn}了下来。它朝哈利\textcolor{blue}{使了一下眼色}。\end{CJK}}\\
\cdashline{1-2}
\multirow[t]{2}{*}{Reference} &
Slowly, very slowly, it \textcolor{blue}{raised} its head until its eyes were \textcolor{blue}{on a level with} Harry's. \\
& It \textcolor{blue}{winked}. \\
\cdashline{1-2}
\multirow[t]{2}{*}{Prediction} & Slowly and very slowly \textcolor{red}{\ul{-- thinking}} his head up, still adding to poster him gladly \textcolor{red}{\ul{stare}} to \\
& stopped Harry's face alone, and then \textcolor{red}{\ul{blurted it out}} to Harry like a stop.\\
\bottomrule
\end{tabular}
\caption{\label{tab:example}Examples with high and low recovery degree, respectively, trained on WMT17 Zh$\Rightarrow$En. We mark errors with \textcolor{red}{\ul{red underline}}.}
\end{table*}

\begin{figure}
    \centering
    \includegraphics[width=0.5\textwidth]{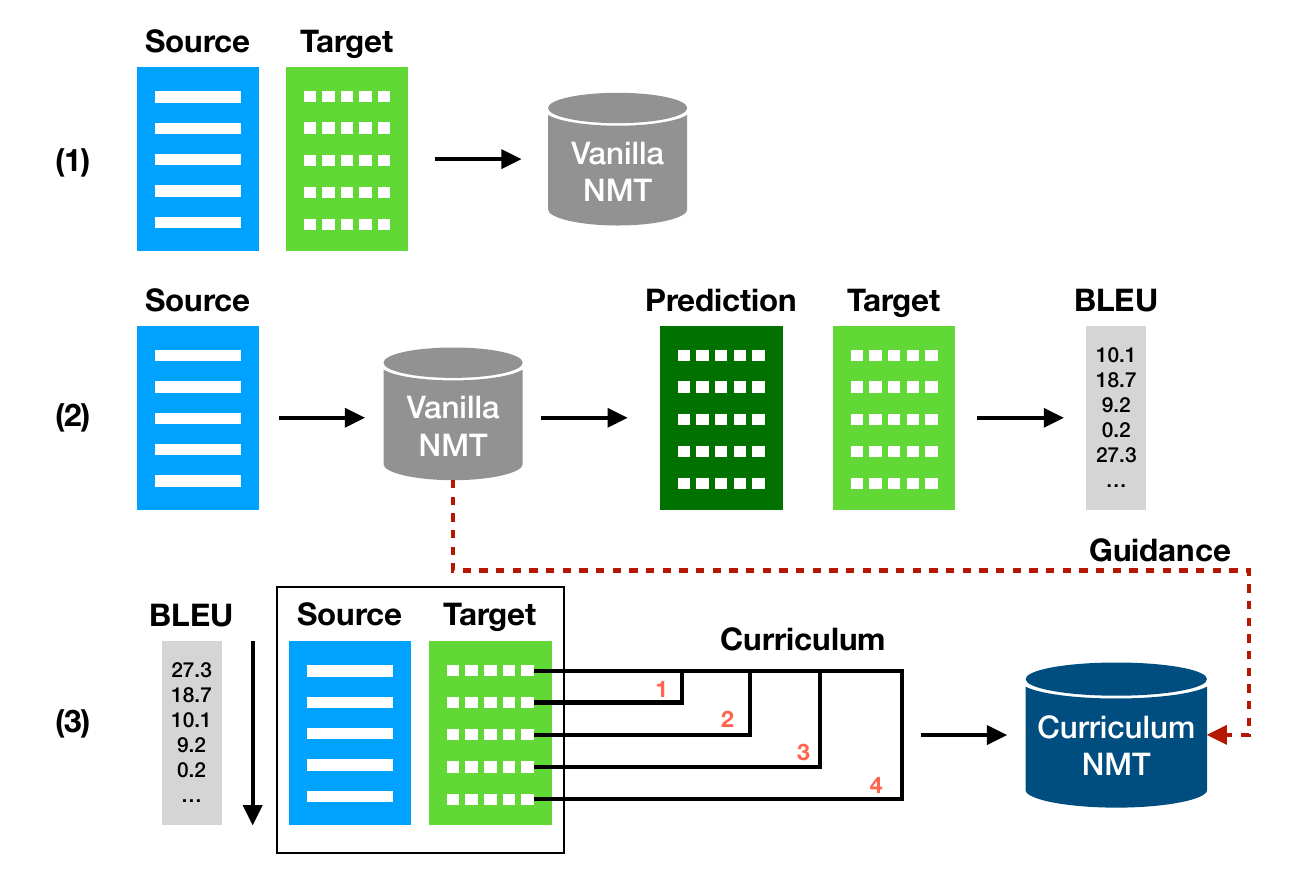}
    \caption{Workflow of self-guided CL strategy}
    \label{fig:workflow}
\end{figure}

\label{sec:method}
We propose a self-guided CL strategy to schedule learning of NMT models with recovery criterion as the learning difficulty. Table~\ref{tab:example} shows two examples with high and low recovery degree, predicted by a well-trained vanilla NMT model. We derive difficulty criterion from this vanilla model and determine curriculum scheduling accordingly. Figure~\ref{fig:workflow} demonstrates the workflow of proposed CL strategy.

\subsection{Difficulty Criterion}
\label{subsec:criterion} 
The loss function of the vanilla model can be written as an average distribution over the training corpus $\mathbb{D}$:

\begin{equation}
\label{eq:cost}
J(\varphi)=\mathbb{E}_{x,y \sim \hat{p}_{\mathbb{D}}} L(f(x^n;\varphi),y^n)
\end{equation}

where $f(x^n;\varphi)$ represent model's prediction and $L$ is the loss function. As noted in Section~\ref{sec:def}, curriculum learning is to minimize the objective $J(\theta)$ with a set of sub-optimal processes from easy to difficult. Examples that better fit into the average distribution learned by the vanilla model with parameter $\varphi$ get higher recovery degree. To start curriculum learning on a set of examples with higher recovery degree is to start optimizing $J(\theta)$ from a smaller parameter space in the neighborhood of parameter $\varphi$. As in machine translation scenario we care more about model performance evaluated by automatic metrics, we choose BLEU score, the de facto metric for MT, to measure the recovery degree. The difficulty criterion based on sentence-level BLEU score is as follows:

\begin{equation}
\label{eq:criterion}
    d(z^n) = -\mathrm{BLEU}(f(x^n;\varphi),y^n)
\end{equation}

Other metrics for MT can also be applied in this difficulty criterion. Based on this criterion, examples with lower difficulty scores are presented at early learning phases, leaving those with higher difficulty scores to the latter phases. 

\subsection{Curriculum Scheduling}
\label{subsec:schedule}

In this paper, we follow the baby steps regimen, in which training corpus is scored and split into subsets before training, as described in Section~\ref{sec:def}. Here we define the corpus splitting function $g$:
\begin{equation}
\label{eq:split}
\begin{aligned}
& g(d(\cdot)):\mathbb{D} \longrightarrow \{\mathbb{D}_1,\dots,\mathbb{D}_K\}, \\
& \mid \forall a \in \mathbb{D}_k, \forall b \in \mathbb{D}_{k+1}, d(a) \leq d(b)
\end{aligned}
\end{equation}
which split training corpus $\mathbb{D}$ into $K$ mutual exclusive subsets $\{\mathbb{D}_1,\dots,\mathbb{D}_K\}$. Each corresponds to a training phase.

On top of this, we explore both fixed and dynamic scheduling settings to train the CL model.

\paragraph{Fixed} The training duration of each training phase is predefined. CL model spends a fixed number of training steps $T$ on each training phase. Subset $\mathbb{D}_k$ is merged into training set at the beginning of the $k$th training phase, see Algorithm~\ref{alg:1}:

\begin{algorithm2e}[t]
\caption{Fixed Scheduling}
\label{alg:1}
\KwIn{Parallel corpus $\mathbb{D}=\{z^n\}_{n=1}^N$, $z^n=(x^n,y^n)$}
Train vanilla model $\varphi$ on $\mathbb{D}$ \\
Compute difficulty score $d(z^n),z^n\in \mathbb{D}$ with $\varphi$ by Eq.~\ref{eq:criterion}\\
Split $\mathbb{D}$ into subsets $\{\mathbb{D}_1,\dots,\mathbb{D}_K\}$ by Eq.~\ref{eq:split} \\
$\mathbb{D}_{\mathrm{train}} = \varnothing$ \\
\For{$k=1,\dots,K$}{
    $\mathbb{D}_{\mathrm{train}} = \mathbb{D}_{\mathrm{train}} \cup \mathbb{D}_{k}$ \\
    \For{training steps $t=1,\dots,T$}{
        Train CL model $\theta^k$ on $\mathbb{D}_{\mathrm{train}}$
        }
    }
\KwOut{Trained CL model $\theta$}
\end{algorithm2e}

\paragraph{Dynamic} We follow the \textit{uncertainty} type of dynamic setting, Section~\ref{sec:def}, in which training duration is controlled by a factor.
We define this factor by model recovery degree. If the CL model constantly demonstrate higher recovery degree than the vanilla model on the newly merged subset $\mathbb{D}_k$ in current training phase $k$ , the CL model training will advance to the learning phase $k+1$. For easier operation, we randomly sub-sample $\mathbb{D}'_{k}$ from $\mathbb{D}_k$. Based on the performance on $\{x^n,y^n\} \in \mathbb{D}'_{k}$, measured by corpus-level BLEU score, we compute model recovery degree of the CL model at current training phase $k$ by:

\begin{equation}
\label{eq:mrec1}
    o_c(k) = \mathrm{BLEU}(f(x^n;\theta^k),y^n)
\end{equation}

We compute model recovery degree of the vanilla model at training phase $k$ with the same subset $\mathbb{D}'_{k}$ by:

\begin{equation}
\label{eq:mrec2}
    o_v(k) = \mathrm{BLEU}(f(x^n;\varphi),y^n)
\end{equation}

If $o_c > o_v$, current training phase $k$ will stop and move to the next. Otherwise, the learning process will remain in the same training phase until it reaches predefined maximum time steps $T$ and then move to the next phase. The training flow is described in Algorithm~\ref{alg:2}.

\begin{algorithm2e}[t]
\caption{Dynamic Scheduling}
\label{alg:2}
\KwIn{Parallel corpus $\mathbb{D}=\{z^n\}_{n=1}^N$, $z^n=(x^n,y^n)$}
Train vanilla model $\varphi$ on $\mathbb{D}$ \\
Compute difficulty score $d(z^n),z^n\in \mathbb{D}$ with $\varphi$ Eq.~\ref{eq:criterion}\\
Split $\mathbb{D}$ into subsets $\{\mathbb{D}_1,\dots,\mathbb{D}_K\}$ by Eq.~\ref{eq:split} \\
$\mathbb{D}_{\mathrm{train}} = \varnothing$ \\
\For{$k=1,\dots,K$}{
    $\mathbb{D}_{\mathrm{train}} = \mathbb{D}_{\mathrm{train}} \cup \mathbb{D}_{k}$ \\
    \For{training steps $t=1,\dots,T$}{
        Train CL model $\theta^k$ on $\mathbb{D}_{\mathrm{train}}$ \\
        Compute model recovery degree $o_c$ and $o_v$, Eq.\ref{eq:mrec1},\ref{eq:mrec2} \\
        \If{$o_c > o_v$}{
            Stop and move to the next phase
        }
    }
}
\KwOut{Trained CL model $\theta$}
\end{algorithm2e}

\section{Experiments}
\label{sec:exp}

\begin{table*}
\centering
\begin{tabular}{c|l||lc|lc|lc}
\toprule
\multirow{2}{*}{\bf \#}&\multirow{2}{*}{\bf Systems}&\multicolumn{2}{c|}{\bf WMT14 EnDe}&\multicolumn{2}{c|}{\bf WMT16 EnDe}&\multicolumn{2}{c}{\bf WMT17 ZhEn}\\
& &BLEU&$\Delta$&BLEU&$\Delta$&BLEU&$\Delta$\\
\hline
\hline
1 & Transformer \textsc{Base} & 27.30 &- & 32.76 &- & 23.69 &- \\
2 & ~~w/ Competence-based CL & 28.19 &- & 32.84 &-& 24.30 &- \\ 
3 & ~~w/ Norm-based CL & 28.81 & -& - &-& 25.25&- \\
4 & ~~w/ Uncertainty-aware CL& - &- &33.93 &-& 25.02&- \\
\hline
\multicolumn{8}{c}{\textit{This work}} \\
\hline
5 & Transformer \textsc{Base} & 27.63 &-& 33.03 &-& 23.78 & - \\
6 & ~~w/ SGCL Fixed & 28.16$^\uparrow$&0.53& 33.55$^\uparrow$ &0.52& 24.65$^\uparrow$&0.87 \\
7 & ~~w/ SGCL Dynamic & \bf 28.62$^\Uparrow$ &0.99& \bf 34.07$^\Uparrow$ &1.04& \bf 25.34$^\Uparrow$&1.56 \\
\bottomrule
\end{tabular}
\caption{\label{tab:result}Experiment results on WMT14 En$\Rightarrow$De with newstest2014 and newstest2016, and WMT17 Zh$\Rightarrow$En comparing with baseline and existing CL methods.``$^{\Uparrow/\uparrow}$'' indicates significant difference ($p < 0.01/ 0.05$) from Transformer \textsc{Base}.} 
\end{table*}

\subsection{Datasets}
We conduct experiments on two machine translation benchmarks: WMT'14 English$\Rightarrow$German (En-De), and WMT'17 Chinese$\Rightarrow$English (Zh-En). 
For En-De, the training set consists of 4.5 million sentence pairs. We use newstest2012 as the validation set and report test results on both newstest2014 and newtest2016 to make better comparison with existing approaches. For Zh-En, we follow~\citep{hassan2018achieving} to extract 20 million sentence pairs as the training set. We use newsdev2017 as the validation set and newstest2017 as the test set. Chinese sentences are segmented with a word segmentation toolkit Jieba\footnote{\url{https://github.com/fxshy/jieba}}. Sentences in other languages are tokenized with Moses\footnote{\url{https://github.com/mosesdecoder}}. We learn Byte-Pair Encoding(BPE)~\citep{sennrich2016neural} with 32k merge operations. And we learn BPE with a shared vocabulary for En-De. We use BLEU~\citep{papineni2002bleu} as the automatic metrics for computing recovery degree and evaluating model performance with statistical significance test~\citep{collins2005clause}.

\subsection{Model Settings}
We conduct experiments with the FairSeq\footnote{\url{https://github.com/pytorch/fairseq}}~\citep{ott2019fairseq} implementation of the Transformer \textsc{Base}~\citep{vaswani2017attention}. For regularization, we use the dropout of 0.2, and label smoothing $\epsilon$ = 0.1. We train the model with batch size of approximately 128K tokens. We use Adam~\citep{kingma2015adam} optimizer and the learning rate warms up to $5\times10^{-4}$ in the first 16K steps, and then decays with the inverse square-root schedule. We evaluate the translation performance on an ensemble of top 5 checkpoints to avoid stochasticity. We use shared embeddings for En-De experiments. All our experiments are conducted with 4 NVIDIA Quadro GV100 GPUs.

\subsection{Curriculum Learning Settings}
The vanilla model and CL model share a same Transformer \textsc{Base} setting. For the recovery degree, we let the trained vanilla model to make predictions of source sentences in the training corpus with beam size set to 1, since at this stage we only need to reveal to which extend a training example can be recovered. Then we evaluate the predictions with BLEU score as the learning difficulty of each training examples. According to~\citet{zhou2020uncertainty}, taking 4 baby steps is superior to other settings, so we also decompose the CL training into 4 training steps. Specifically, we find it helpful if model training warm-up every time when a new subset is merged into the training set. Based on the proposed difficulty criterion, we investigate two curriculum scheduling methods:

\begin{itemize}
    \item \textbf{SGCL Fixed} represents self-guided curriculum learning with fixed scheduling.
    \item \textbf{SGCL Dynamic} represents self-guided curriculum learning with dynamic scheduling.
\end{itemize}

\section{Results}
\label{sec:results}

Table~\ref{tab:result} summarises our experimental results together with that of recent curriculum learning methods. Row 1 are the results of standard Transformer \textsc{Base} baseline of these benchmarks and row 2-4 demonstrate existing curriculum learning approaches. Row 5 are the results of our own Transformer \textsc{Base} implementation, and 6-7 are that of our proposed curriculum learning method. For En-De benchmark, if existing curriculum learning approaches only report results on one of the newstest2014 or newstest2017, then only the reported one is shown. We report results on them both for better comparison.

We implement Transformer \textsc{Base} with 300k training steps for baseline and proposed CL methods. For both SGCL Fixed and SGCL Dynamic methods, we observe superior performances over the strong baseline on all three test sets of two benchmarks, which agree with existing approaches that curriculum learning can facilitate NMT model. And if we compare two scheduling methods, SGCL Dynamic outperforms SGCL Fixed. A possible reason is that dynamic CL scheduling encourages CL model to spend more steps on more difficult subset. Encouragingly, we observe considerable gains over other curriculum learning counterparts.

\section{Analysis}
\label{sec:analysis}

\subsection{Recovery Degree}

\begin{figure}
    \centering
    \includegraphics[width=0.35\textwidth]{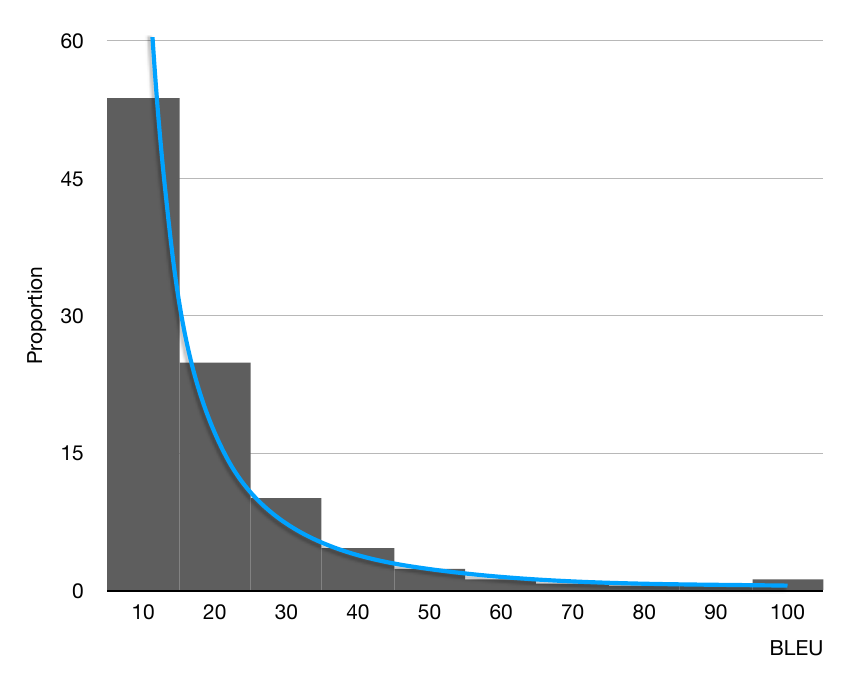}
    \caption{\label{fig:proportion}Recovery degree (BLEU) distribution of the training corpus.}
\end{figure}

We conduct experiments on En-De benchmark for further analysis of proposed curriculum learning methods.

As described in Section~\ref{sec:method}, we adopt metric BLEU to measure the recovery degrees of all examples in the training corpus with the pre-trained vanilla model. When making predictions with the vanilla model, the beam size is set to 1 for simplicity. So the recovery degrees measured by BLEU score could be lower than test results of strong baseline. If we look at the distribution in terms of BLEU score on all training examples, as Figure~\ref{fig:proportion} illustrated, the distribution is very dense in the range within lower scores. Specifically more than 53.9\% training examples get a recovery degree lower than 10. This indicates that for a well-trained vanilla model, the empirical distribution and the model distribution are still inconsistent on some examples. According to our proposed difficulty criterion and curriculum scheduling, the training corpus is split into 4 subsets with equal size, $\{\mathbb{D}_1,\mathbb{D}_2,\mathbb{D}_3,\mathbb{D}_4\}$ that are merged aggressively during training. Table~\ref{tab:bleu} is the range and average of recovery degrees of each subset, revealing the increasing of learning difficulty as training phases progresses.

\begin{table}
    \centering
    \begin{tabular}{ccr}
        \toprule
        Subset & {Range} & Average\\
        \hline
        $\mathbb{D}_1$ & 17.72 - 100.00 & 35.62 \\
        
        $\mathbb{D}_2$ & 9.18 - 17.72 & 12.77 \\
        
        $\mathbb{D}_3$ & 5.16 - 9.18 & 6.97 \\
        
        $\mathbb{D}_4$ & 0.00 - 5.16 & 3.35 \\
        \bottomrule
    \end{tabular}
    \caption{\label{tab:bleu}Range of recovery degree (BLEU) in subsets $\{\mathbb{D}_1,\mathbb{D}_2,\mathbb{D}_3,\mathbb{D}_4\}$}
\end{table}

\subsection{Learning Curves}

\begin{figure}
\centering
    \subfloat[Baseline vs. SGCL Fixed]{
        \includegraphics[width=0.45\textwidth]{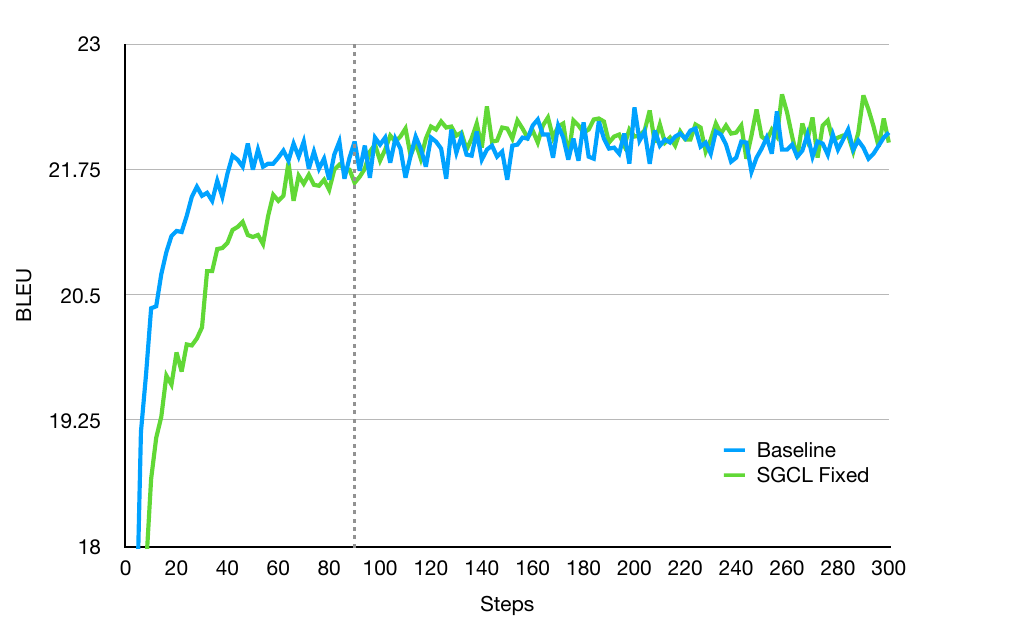}} \\
    \subfloat[Baseline vs. SGCL Dynamic]{
        \includegraphics[width=0.45\textwidth]{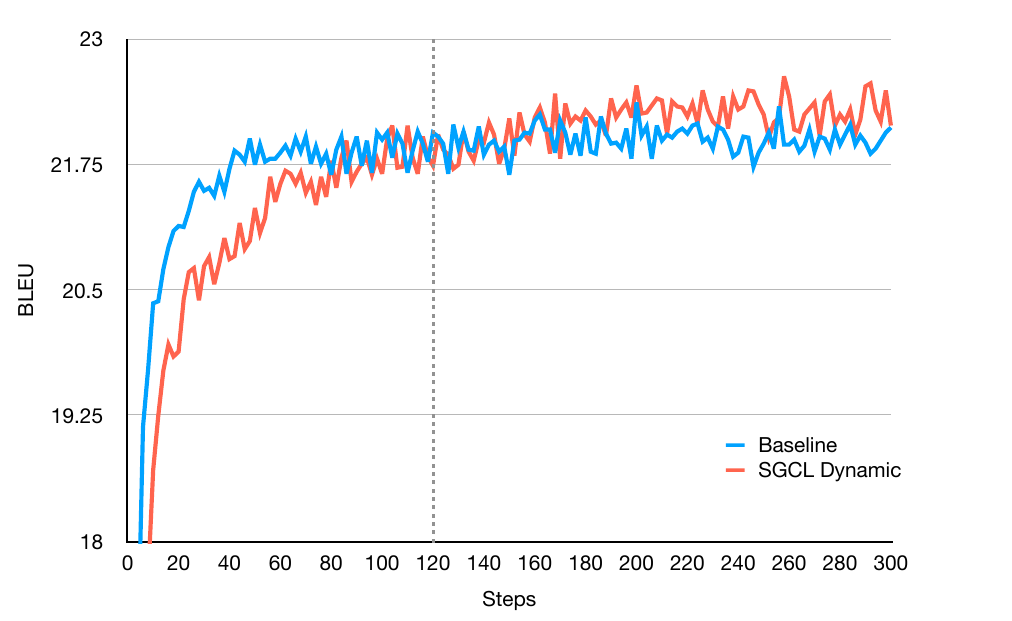}}
    \caption{\label{fig:clcurve}Learning curves w.r.t BLEU scores.}
\end{figure}

Figure~\ref{fig:clcurve} demonstrates the learning curves of baseline vs. SGCL Fixed and baseline vs. SGCL Dynamic. As illustrated, baseline converges faster at the beginning but stays at a lower level as training progresses, while proposed CL methods can gain constant improvements and outperforms the baseline in later training process. A possible reason that the CL model doesn't outperform the baseline at the beginning might be it boosts the performance after all training examples are merged into the training set. After all training examples are included, CL models are able to maintain better growth momentum than the baseline.

We also observe that the SGCL Dynamic gains more significant improvements over the baseline than the SGCL Fixed. Considering a total of 300k training steps, different curriculum scheduling is actually different ways of splitting the training steps. For the SGCL Fixed, we empirically define the training steps spend on phase 1 to phase 4 as 30k, 30k, 30k, 210k. That is to say, after 90k steps, the model is training with all examples in the training corpus. For SGCL Dynamic, as mentioned in Section~\ref{sec:method}, if the CL model outperforms the vanilla model on newly merged subset, training progresses to the next phase. In practice, after new examples merged into training set, we first train for 20k steps and then check the performance of the CL model every 10k steps. If the CL model outperforms the vanilla model successively, training will move to the next phase. As a result, the model starts to train with all training examples after 120k steps, and tends to spend more time steps in latter training phases, which is consistent with other existing dynamic scheduling methods. 

\subsection{Case Study}
Figure~\ref{tab:case} presents a case study on Zh-En. It indicates that our approach achieves performance boost because of better lexical choice, which is more close to the reference sentence. To better understand how our approach alleviates the low-recovery problem, we conduct statistic analysis on the BLEU scores of predictions made by the vanilla model and the CL model on test set. Testing results show that the proportion of predictions with BLEU score under 10 is $10.0\%$ with the vanilla model, and is down to $8.1\%$ with the CL one.

\begin{table*}
\renewcommand\arraystretch{1.2}
\centering
\begin{tabular}{ll}
\toprule
\multirow[t]{2}{*}{Source} & {\begin{CJK}{UTF8}{gbsn}然而, 就在大部分互联网医疗企业挣扎在A轮或B轮的\textcolor{blue}{融资}路上的时候,\end{CJK}}\\ & {\begin{CJK}{UTF8}{gbsn}有几家\textcolor{blue}{细分领域领先企业}仍能获得资本热捧。\end{CJK}} \\
\cdashline{1-2}
\multirow[t]{3}{*}{Reference} & However, just as the majority of internet medical companies struggle on the way of \\ 
& a round or b round of \textcolor{blue}{financing}, several \textcolor{blue}{segment-leading enterprises} can still be \\
& favored by investors.\\
\cdashline{1-2}
Vanilla & However, even as most internet healthcare companies struggle to \textcolor{red}{\ul{raise money}} in a or \\ 
(8.61) & b rounds, a few of the \textcolor{red}{\ul{leading segments}} still enjoy the capital boom. \\
\cdashline{1-2}
SGCL & However, even as most internet health companies struggle with a round or b round of \\
(27.45) & \textcolor{blue}{financing}, several \textcolor{blue}{segments leading} \textcolor{red}{\ul{business}} still enjoy the capital boom. \\
\bottomrule
\end{tabular}
\caption{\label{tab:case} Prediction made by Vanilla model and SGCL Dynamic model with the same input sentence from test set. We mark errors with \textcolor{red}{\ul{red underline}}. And the number in parentheses, e.g. (8.61) is sentence level BLEU score.} 
\end{table*}

\section{Conclusion}
\label{sec:conclusion}
In this work, we propose a self-guided CL strategy for neural machine translation. The intuition behind it is that after skimming through all training examples, the NMT model naturally learns how to schedule a curriculum for itself. We then manage to discuss existing difficulty criteria for curriculum learning from a probabilistic perspective, which also explains our motivation of deriving difficulty criterion based on the idea of recovery. Moreover, we corporate this recovery based difficulty criterion with both fixed and dynamic curriculum scheduling. Empirical results show that with self-guided CL strategy, NMT model achieves better performance over the strong baseline on translation benchmarks and our dynamic scheduling outperforms the fixed one. 
In the future, we will corporate recovery based difficulty criterion with other dynamic scheduling methods. Also, it will be interesting to apply our CL strategy on other scenarios, e.g. non-autoregressive generation~\citep{Gu2018NonAutoregressiveNM,wu-etal-2020-slotrefine,ding-etal-2020-context}.



\bibliographystyle{acl_natbib}
\bibliography{reference}

\end{document}